\documentclass[10pt,twocolumn,letterpaper]{article}

\usepackage[pagenumbers]{iccv}
\usepackage{multirow}
\usepackage{amsmath}
\usepackage{xcolor}
\usepackage{lineno}
\usepackage{float}
\usepackage{stfloats}
\definecolor{darkgrey}{rgb}{0.66, 0.66, 0.66}
\DeclareMathOperator*{\argmin}{arg\,min}

\definecolor{iccvblue}{rgb}{0.21,0.49,0.74}
\usepackage[pagebackref,breaklinks,colorlinks,allcolors=iccvblue]{hyperref}

\newcommand{\method}{{\sc CameraCtrl\:II}\xspace}

\newcommand{\dataset}{{\sc RealCam}\xspace}

\title{CameraCtrl II: Dynamic Scene Exploration via \\ Camera-controlled Video Diffusion Models}

\author{
Hao He$^{1,2}$~~~Ceyuan Yang$^{2,\dag}$~~~Shanchuan Lin$^{2}$~~~Yinghao Xu$^{3}$~~~Meng Wei$^{4}$~~~Liangke Gui$^{2}$ \\
Qi Zhao$^{2}$~~~Gordon Wetzstein$^{3}$~~~Lu Jiang$^{2}$~~~Hongsheng Li$^{1}$ \\
$^{1}$The Chinese University of Hong Kong~~~$^{2}$ByteDance Seed~~~$^{3}$Stanford University ~~~$^{4}$ByteDance \\
{\small $\dag$ corresponding author}\\
{\tt\small \url{https://hehao13.github.io/Projects-CameraCtrl-II/}}
}

\begin{document}
\twocolumn[{
\renewcommand\twocolumn[1][]{#1}
\maketitle
\begin{center}
    \centering
    \captionsetup{type=figure}
    \includegraphics[width=1.0\textwidth]{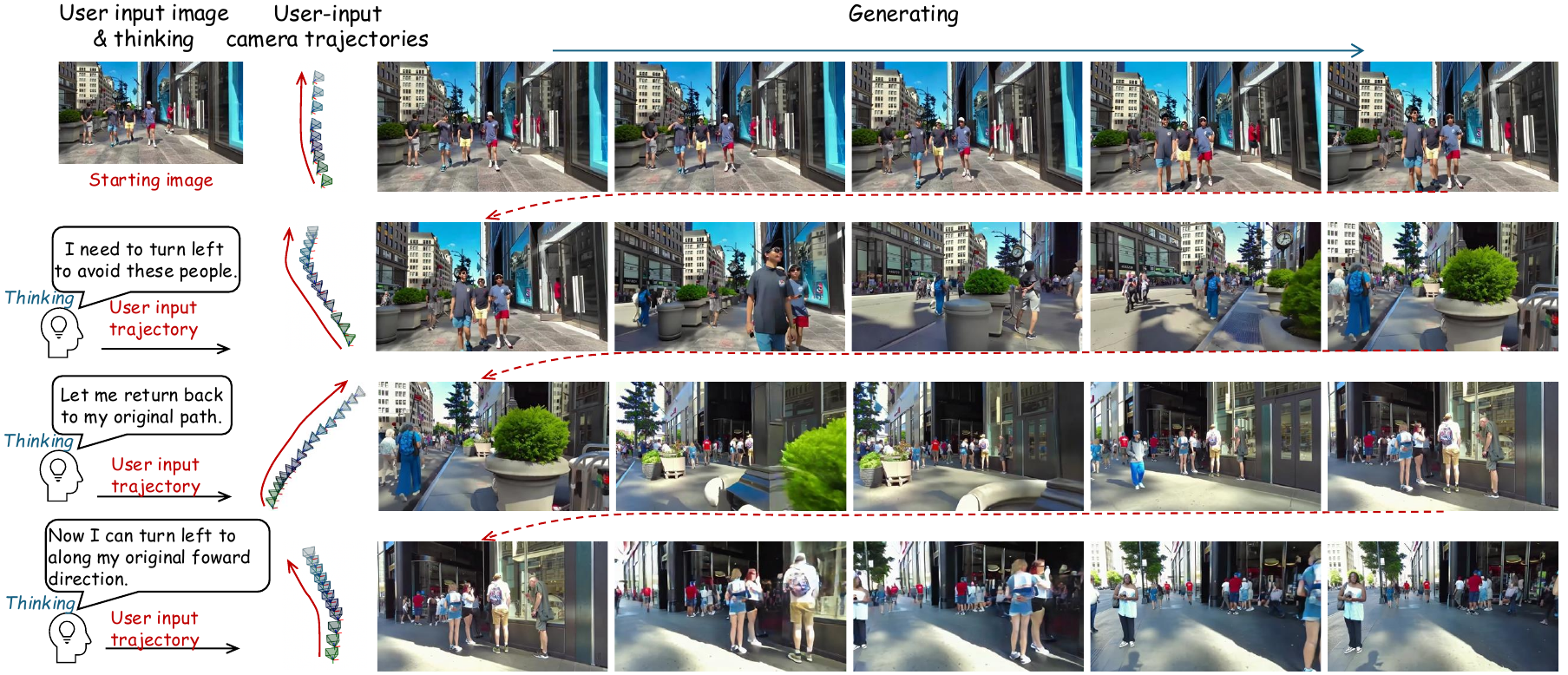}
    \caption{
        \textbf{Illustration of \method.}
        Our camera-controlled video diffusion model generates consistent video sequences for dynamic scenes based on user-defined camera trajectories.
        The first row represents a generated video clip conditioned on the starting image and a user input camera trajectory. 
        After watching the generated video clip, user can decide next step and specify the corresponding camera trajectories.
        Subsequent rows show clips conditioned on previous generated videos and these newly provided camera trajectories. 
        The model strictly follows these user camera trajectory inputs while maintaining scene consistency across multiple video clips, enabling seamless navigation around pedestrians and exploration of the environment from various perspectives.
        }
    \label{fig:teaser}
\end{center}
}]

\begin{abstract}
This paper introduces \method, a framework that enables large-scale dynamic scene exploration through a camera-controlled video diffusion model. 
Previous camera-conditioned video generative models suffer from diminished video dynamics and limited range of viewpoints when generating videos with large camera movement.
We take an approach that progressively expands the generation of dynamic scenes---first enhancing dynamic content within individual video clip, then extending this capability to create seamless explorations across broad viewpoint ranges.
Specifically, we construct a dataset featuring a large degree of dynamics with camera parameter annotations for training while designing a lightweight camera injection module and training scheme to preserve dynamics of the pretrained models. 
Building on these improved single-clip techniques, we enable extended scene exploration by allowing users to iteratively specify camera trajectories for generating coherent video sequences.
Experiments across diverse scenarios demonstrate that \method enables camera-controlled dynamic scene synthesis with substantially wider spatial exploration than previous approaches.

\end{abstract}

\section{Introduction}
Recent years have witnessed remarkable advances in video diffusion models~\cite{svd,sora,kling}, which can generate high-fidelity and temporally coherent videos from text descriptions.
These models accept user-defined control~\cite{sparsectrl,xiaoyu_motioni2v,cameractrl,fu20243dtrajmaster} and are also scalable w.r.t. dataset size and computational resources, producing long and physically plausible videos.
For example, Sora~\cite{sora} can generate minute-long videos with realistic physics and complex motions.
Therefore, these video diffusion models have become a promising tool for modeling and simulating dynamic real-world scenes.

Beyond generating individual dynamic scenes, enabling users to actively explore these digital worlds has become increasingly important.
Recent works have made progress in learning to explore generated spaces.
In the domain of game generation, methods like~\cite{yu2025gamefactory,thematrix,parkerholder2024genie2,gamengen,gamengen} learn to simulate state transitions and predict future observations from action sequences such as keyboard inputs.
For general video generation, camera control has emerged as a natural interface for scene exploration.
Recent works~\cite{cameractrl, camco, ac3d,motionctrl,bahmani2024vd3d,kuang2025collaborative,cavia,ac3d} inject camera parameters into pretrained video diffusion models to enable precise camera viewpoint manipulation.
By controlling virtual camera movements within these generated environments---analogous to camera navigation in the real world---users can explore these generated digital scenes from various perspectives.

Despite their effectiveness in camera control and achieving the exploration in certain spatial range, existing methods face two limitations that hinder their practical applications. 
First, after incorporating camera control, these models often suffer from significant degradation in generating dynamic content.
Second, they are restricted to generating short video clips (e.g., 25 frames for CameraCtrl~\cite{svd}, 49 frames for AC3D~\cite{ac3d}) and cannot generate new clips in the same scene based on previously generated content and new camera trajectories given by users.
These limitations fundamentally limit both the types of scenes that can be generated (constrained to largely static content) and the spatial range that can be explored, thus significantly diminishing the user experience.
We introduce \method to address these two limitations.

We address the challenge of generating highly dynamic videos using two key techniques.
First, existing approaches primarily rely on static video datasets with camera parameter annotations, such as RealEstate10K~\cite{realestate10k} and DL3DV10K~\cite{ling2024dl3dv}.
Training on these datasets inevitably compromises the dynamic capabilities of camera-controlled video diffusion models.
Therefore, we construct a new dataset by extracting camera trajectory annotations from real dynamic videos using Structure-from-Motion (SfM), specifically VGGSfM~\cite{wang2024vggsfm}.
Then, we propose methods to address challenges of arbitrary scale and long-tailed camera trajectory distributions in the constructed dataset.
Second, for the model architecture, we inject camera parameters only at the initial layer of the diffusion model, avoiding over-constraining pixel generation to preserve dynamic content.
Besides, we jointly train all model parameters on both labeled and unlabeled videos to preserve the pretrained model's capability for generating dynamic and diverse scenes while maintaining its ability to perform general video generation tasks, such as text-to-video generation without camera input.
This strategy also enables camera classifier-free guidance~\cite{cfg} during inference to enhance camera control accuracy.

To enable exploration of broader scene ranges, we carefully design a video extension scheme and corresponding training strategy that allows our model to generate multiple coherent video clips sequentially.
Specifically, we extend our single-clip camera-controlled video diffusion model to support clip-wise autoregressive video generation with a novel technique.
During training, the model learns to generate new clips by conditioning on clean frames from previous clips and new camera trajectories, while optimizing only on the newly generated frames.
At inference time, the model can generate new video segments by conditioning on both the previous clip frames and new camera trajectories, maintaining visual consistency while following user desired camera path.
To preserve the high-quality generation capability of single clips, we jointly train this video extension task with the original single-clip camera-controlled video generation. 
~\cref{fig:teaser} shows an example of our CameraCtrl II model generating multiple video sequences in a dynamic urban scene.
It maintains consistent scene content while strictly adhering to camera controls, allowing for diverse exploration patterns such as navigating around pedestrians or changing direction while preserving object motion.

In summary, our key contributions include: 1) A systematic data curation pipeline for constructing a dynamic video dataset with camera trajectory annotations; 2) A lightweight camera control injection module and corresponding training strategy that preserves dynamic video generation capabilities while adding camera control effect; 3) A clip-wise autoregressive generation recipe that enables extended range exploration of generated scenes.

\section{Related Work}
\label{sec:related_work}

\noindent \textbf{Video Diffusion Models.}
Driven by the advances in model architectures~\cite{ho2022vdm}, large-scale datasets~\cite{webvid10m,panda70m,HDVILA100M}, comprehensive benchmarks~\cite{vbench,vbench++}, and improved training techniques~\cite{edm, recitfiedflow}, the field of video diffusion models has seen remarkable progress in recent years.
A major focus of this field has been text-to-video (T2V) generation~\cite{alignyourlatents, cogvideo, makeavideo, svd, lin2024open, ma2024latte,sora, yang2024cogvideox,kong2024hunyuanvideo}, where models create videos from text descriptions. 
Early works~\cite{alignyourlatents,svd,guo2023animatediff,gupta2024photorealistic,cogvideo} efficiently transform UNet-based text-to-image (T2I) models into video generators by incorporating additional temporal modeling layers. 
%
Recent models~\cite{sora,kong2024hunyuanvideo,kling,yang2024cogvideox,noauthor_cosmos_nodate,ma2025step} adopt transformer architectures~\cite{dit,sd3} to achieve better temporal consistency and generation quality at scale.
While these works focus on pretraining models for general-purpose video generation, our work aims to leverage video diffusion models for dynamic scene exploration.
Through video generation, we enable users to freely explore a dynamic scene in a large range.

\noindent \textbf{Camera-controlled Video Diffusion Models.}
To enable camera pose control in the video generation process,  MotionCtrl~\cite{motionctrl}, CameraCtrl~\cite{cameractrl}, I2VControl-Camera~\cite{i2vcontrolcamera} inject the camera parameters~(extrinsic, Plücker embedding~\cite{lfn}, or point trajectory) into a pretrained video diffusion model.
Building upon this, CamCo~\cite{camco} integrates epipolar constraints into attention layers, while CamTrol~\cite{camtrol} leverages explicit 3D point cloud representations.
AC3D~\cite{ac3d} carefully design the camera representation injection to the pretrained model.
VD3D~\cite{bahmani2024vd3d} enables camera control to transformer-based video diffusion models~\cite{snapvideo}.
Several recent works have advanced beyond single-camera scenarios: CVD~\cite{kuang2025collaborative}, Caiva~\cite{cavia}, Vivid-ZOO~\cite{vividzoo}, and SyncCamMaster~\cite{bai2024syncammaster} have developed frameworks for multi-camera synchronization. 
Despite these advances, existing methods struggle to generate dynamic content with camera control, and are limited to short video clips.
Our work enhances dynamic content generation and enables scene exploration through sequential video generation.

\section{CameraCtrl II} \label{sec:method}
We present \method to enable camera-controlled generation of large-scale dynamic scenes using video diffusion model.
To generate such a video with a high degree of dynamism, we carefully curate a new dataset (\cref{subsec:data}) and develop an effective camera control injection mechanism (\cref{subsec:model}).
\cref{subsec:v2v_ext} presents our approach to enable large range exploration in the dynamic scene via a video extension technique. 
\cref{subsec:preliminary} provides essential preliminary on camera-controlled video diffusion models.

\subsection{Preliminary} \label{subsec:preliminary}
Given a pre-trained latent video diffusion model and camera representation $s$, a camera-controlled video diffusion model learns to model the conditional distribution $p(z_0|c, s)$ of video tokens, where $z_0$ represents the encoded latents from a visual tokenizer~\cite{magvit2} and $c$ denotes the text/image prompt.
The training process involves adding noise $\epsilon_t$ to the latents at each timestep $t\in[0, T]$ to obtain $z_t$ and optimizing a transformer model to predict this noise using the following objective:
\begin{equation}
L(\theta) = \mathbb{E}_{z_0, \epsilon, c, s, t}[|\epsilon - \hat{\epsilon}_{\theta}(z_t, c, s, t)|^2_2].
\label{equ:diffusion_obj_control}
\end{equation}
For inference, we initialize from Gaussian noise $\epsilon \sim \mathcal{N}(0, \sigma_t^2\mathbf{I})$ and iteratively recover the video latents $z_0$ using the Euler sampler, conditioning on both the input image and camera parameters.

For the camera representation, we follow recent works~\cite{cameractrl, camco} and adopt the Plücker embedding~\cite{lfn}, which provides strong geometric interpretation and fine-grained per-pixel camera information.
Specifically, given camera extrinsic matrix $\mathbf{E}=\lbrack \mathbf{R}; \mathbf{t} \rbrack \in \mathbb{R}^{3 \times 4}$ and intrinsic matrix $\mathbf{K} \in \mathbb{R}^{3 \times 3}$, we compute for each pixel $(u, v)$ its Plücker embedding $\mathbf{p} = (\mathbf{o} \times \mathbf{d}', \mathbf{d}')$.
Here, $\textbf{o}$ represents the camera center in world space, $\mathbf{d} = \mathbf{RK^{-1}}[u, v, 1]^T + \mathbf{t}$ denotes the ray direction from camera to pixel, and $\textbf{d}'$ is the normalized $\textbf{d}$.
The final Plücker embedding $\mathbf{P}_i \in \mathbb{R}^{6 \times h \times w}$ is constructed for each frame, with spatial dimensions $h$ and $w$ matching those of the encoded visual tokens.

\begin{figure}[t]
	\centering
	\includegraphics[width=\linewidth]{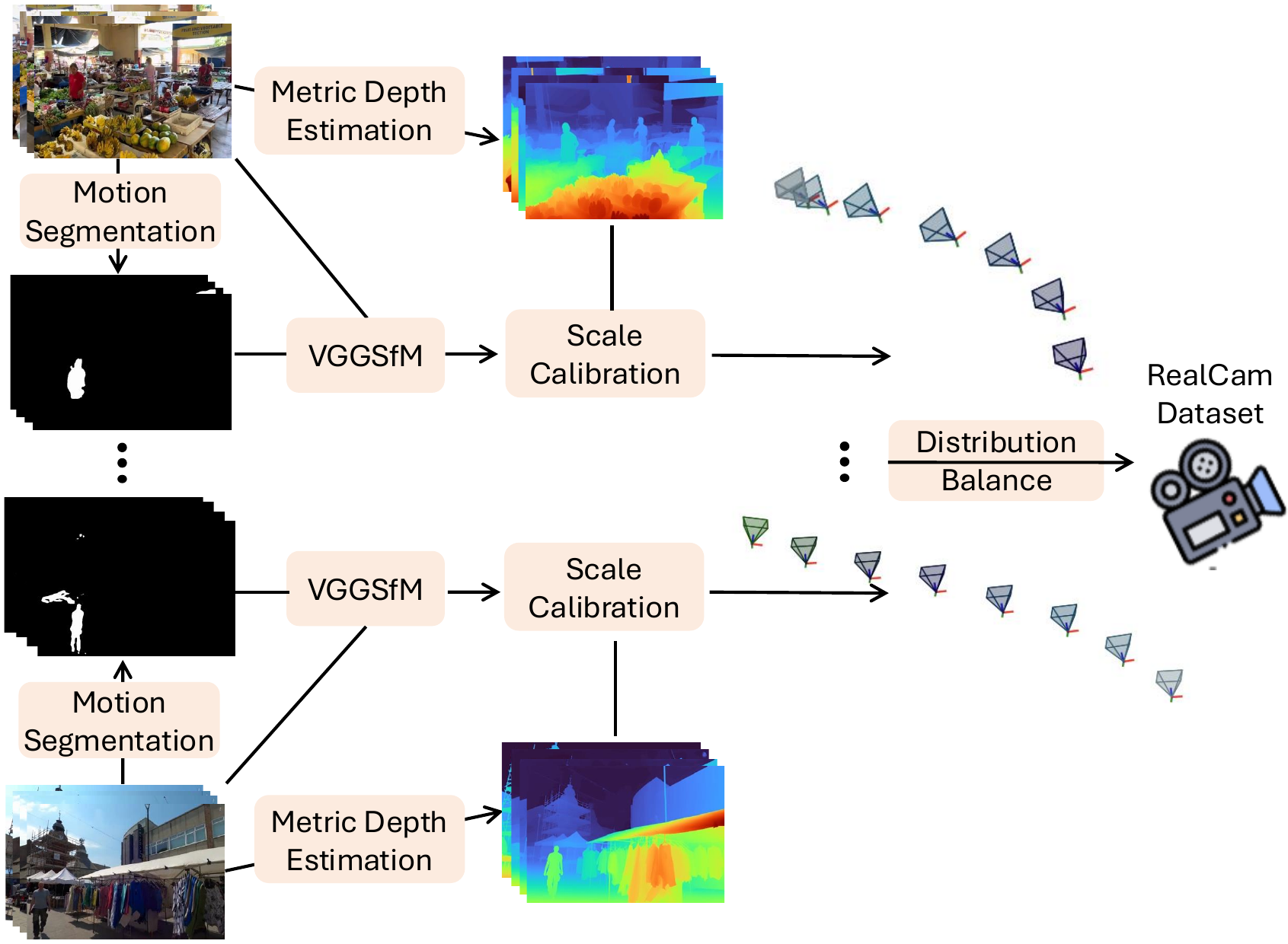}
	\caption{\textbf{Dataset curation pipeline.} We omit the process of dynamic video selection.}
	\label{fig: datapipeline}
    \vspace{-1.6em}
\end{figure}

\subsection{Dataset Curation} \label{subsec:data}
High-quality video datasets with accurate camera parameter annotations are crucial for training a camera-controllable video diffusion model. 
While existing datasets, like RealEstate10K~\cite{realestate10k}, ACID~\cite{infinite_nature_2020}, DL3DV10K~\cite{ling2024dl3dv} and Objaverse~\cite{objaverse} provide diverse camera parameter annotations, they primarily contain static scenes and focus on single domain.
Previous works such as CameraCtrl~\cite{camco}, MotionCtrl~\cite{motionctrl}, and Camco~\cite{camco} have shown that training on these static scene datasets leads to significant degradation in dynamic content generation.
To address this limitation, we introduce \dataset Dataset, a new dynamic video dataset with precise camera parameter annotations.
The overall data processing pipeline is shown in~\cref{fig: datapipeline}.

\begin{figure*}[t]
	\centering
	\includegraphics[width=\linewidth]{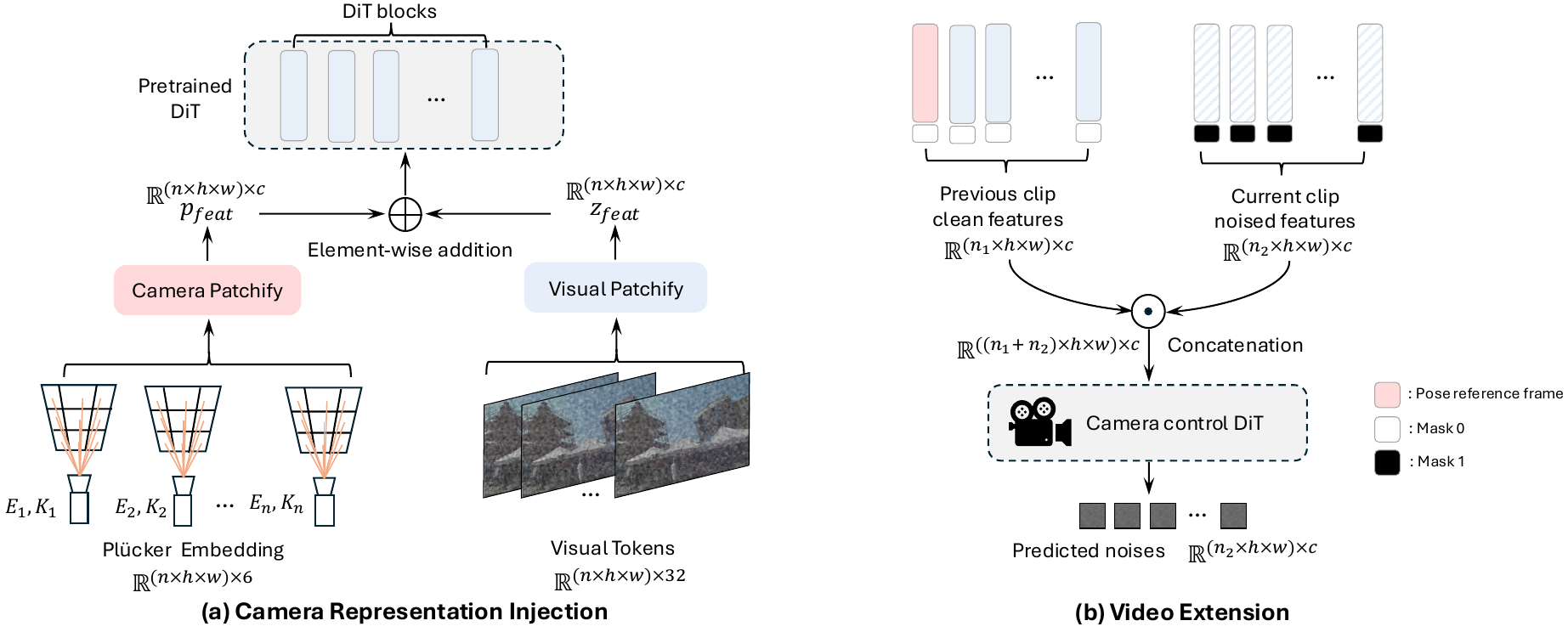}
	\caption{
        \textbf{Model architecture of \method.} (a) Given a pretrained video diffusion model, \method adds an extra camera patchify layer at the initial of the model.
        It takes the Plücker embedding as input, outputs camera features with the same shape of the visual features.
        Both features are element-wisely added before the first DiT layer.
        (b) Features belonging to the previous video clip are kept clean, while current features are noised.
        After concatenation, features are sent to a camera control DiT; we only compute the loss of the current clip's tokens.
        We omit the text encoder for both figures, and the camera features for the second figure.
        }
	\label{fig: modelarch}
\end{figure*}

\noindent\textbf{Camera Estimation from Dynamic Videos.} While synthetic scenes used in recent works~\cite{fu20243dtrajmaster,bai2024syncammaster} could provide precise camera parameter annotation, they require extensive manual design of individual scenes and environments.
This labor-intensive process significantly limits dataset scalability and diversity.
Therefore, we opt to curate our dataset from real-world videos.
To maintain scene diversity, we get videos across various scenarios including indoor environments, aerial views, and street scenes.
Our data processing pipeline consists of several key steps: 
First, we employ the motion segmentation model TMO~\cite{tmo} to identify dynamic foreground objects in a video.
Then RAFT~\cite{raft} is used to estimate optical flow of a video.
With the mask and the optical flow, by averaging the optical flow in static background regions, we obtain a quantitative measure of camera movement.
Videos are selected only when their average flow exceeds an empirically determined threshold, ensuring sufficient camera movement.
After that, we use VGGSfM~\cite{wang2024vggsfm} to estimate camera parameters for each frame. 
However, initial experiments revealed two key challenges: 1) Structure-from-Motion reconstructions from monocular videos inherently produce arbitrary scene scales, making it difficult to learn consistent camera movements.
2) Real-world videos have imbalanced camera trajectory distribution, with certain camera trajectory types like forward motion being overrepresented.
This can cause the model to overfit to common trajectory types while performing poorly on underrepresented types of camera movements.
Therefore, we do the following two modification of our dataset.

\noindent\textbf{Camera Parameter Calibration for Unified Scales.} To establish a unified scale across scenes, we develop a calibration pipeline aligning arbitrary scene scales to metric space. For each video sequence, we first select $N$ keyframes and estimate their metric depths $\{\mathbf{M}_i\}^N_{i=1}$ using a metric depth estimator~\cite{depthpro}.
We then obtain corresponding SfM depths $\{\mathbf{S}_i\}^N_{i=1}$ from the VGGSfM output.
The scale factor $s$ between metric and VGGSfM depths for each frame $i$ can be formulated as:
\begin{equation}\label{equ:ransac}
s_i = \argmin_s \sum_{p \in \mathcal{P}} \rho(|s \cdot \mathbf{S}_i(p) - \mathbf{M}_i(p)|)
\end{equation}
where $\mathcal{P}$ denotes pixel coordinates, and $\rho(\cdot)$ is the Huber loss function.
We solve this minimization problem using RANSAC~\cite{mvg} to ensure robustness against depth estimation errors.
The final scale factor $s$ for the a scene is computed as the mean of individual frame scales.
This factor is multiplied to the camera position vector $\mathbf{t} \in\mathbb{R}^{3 \times 1}$ of the extrinsic matrix, obtaining $\mathbf{E}=\lbrack \mathbf{R}; s\cdot\mathbf{t} \rbrack \in \mathbb{R}^{3 \times 4}$.

\noindent\textbf{Camera Trajectory Distribution Balancing.} We implement a systematic approach to analyze, and balance the distribution of camera trajectory types.
We first detect key camera positions~(keypoints) on a camera trajectory: for each point, we fit two lines through its preceding and following $n$ points, marking it as a keypoint if the angle between these lines exceeds threshold $\gamma$.
These keypoints divide a camera trajectory into several segments, whose directions are determined by the fitted line vectors.
The segment with the longest camera movement defines the trajectory's primary movement direction.
Along each segment, we analyze camera rotation matrices to identify significant view changes. 
Between adjacent segments, we identify turns by measuring their angular deviations, with turns after the main segment defined as the main turns of the trajectory.
Each trajectory is assigned an importance weight based on the number and magnitude of both view changes and turns.
We then categorize trajectories into $N \times M$ categories based on N primary directions and M main turns.
To balance the dataset, we prune redundant trajectory types by removing trajectories with lower importance scores, resulting a more uniform camera trajectory distribution of the dataset.

\subsection{Adding Camera Control to Video Generation} 
\label{subsec:model}
With our dataset of dynamic videos and corresponding camera parameter annotations in hand, we next explore how to enable camera control in video diffusion models and preserve the dynamics of generated videos.
This requires careful design of both the camera parameter injection module and training strategies.
A key challenge is to incorporate camera control while maintaining the model's ability to generate dynamic scenes.
We detail our approach in the following sections.

\noindent \textbf{Lightweight Camera Injection Module.}
Previous methods~\cite{cameractrl, camco, motionctrl, fu20243dtrajmaster,bai2024syncammaster} often employ a dedicated encoder to extract camera features and then inject them into each diffusion transformer~(DiT) or convolution layers.
These global camera injection approaches~\cite{ac3d} can over-constrain video dynamics, limiting natural motion variations in the generated content.
Instead, we inject camera condition only at the initial layer of diffusion models using a new patchify layer for camera tokenization that matches the dimensions and downsample ratios of visual patchify layers.
The visual tokens $z_t$ and Plücker embeddings $p$ are processed through their respective patchify layers to get visual features $z_{feat}$ and camera features $p_{feat}$. 
And they are combined via element-wise addition ($z_{feat}=z_{feat} + p_{feat}$) before flowing through remaining DiT layers, as shown in~\cref{fig: modelarch} (a).
This simple yet effective approach preserves dynamic motion better than encoder-injector methods while achieving superior camera control, as demonstrated in \cref{tab:ablation_model_arch}.

\noindent \textbf{Joint Training with Camera-labeled and Unlabeled data.}
Training the DiT model on \dataset with improved camera injection modules still limits the model's capability to generate diverse content since \dataset only covers a subset of scenes compared to the pretraining data.
To address this limitation, we propose a joint training strategy that leverages both camera-labeled and unlabeled video data.
For labeled data, we incoperate Plücker embeddings from the estimated camera parameters as previously described.
For unlabeled data, we utilize an all-zero dummy Plücker embedding as the condition input.
This joint training framework enables an additional advantage: implementing classifier-free guidance~(cfg) for camera control, analogous to widely-adopted classifier-free text guidance~\cite{cfg}.
We formulate camera classifier-free guidance as:
\begin{align}
    \hat{\epsilon}_\theta(z_t, c, s, t) &= \epsilon_\theta(z_t, \phi_{text}, \phi_{cam}) \nonumber \\
    &+ w_{text}(\epsilon_\theta(z_t, c, \phi_{cam}) - \epsilon_\theta(z_t, \phi_{text}, \phi_{cam})) \nonumber \\
    &+ w_{cam}(\epsilon_\theta(z_t, c, s) - \epsilon_\theta(z_t, c, \phi_{cam}))
\end{align}
where $z_t$ denotes the noised latent at timestep $t$, $\epsilon_\theta$ represents the denoising network, $\phi$ indicates null conditioning, and $w_{text}$ and $w_{cam}$ are guidance weights for text and camera conditions respectively.
This formulation allows enhancing camera control accuracy through appropriate adjustment of guidance weights.
With this training scheme, our model learns effective camera conditioning while maintaining good generalization to in-the-wild senarios.

\subsection{Sequential Video Generation for Scene Exploration}  \label{subsec:v2v_ext}
After obtaining a model capable of generating camera-controlled dynamic video, we try to enable broader scene exploration through sequential video generation.

\noindent\textbf{Clip-level Video Extension for Scene Exploration.} 
We extend our single-clip camera-controlled video diffusion model to support clip-wise sequential generation.
During training, for a previously generated video clip $i$, we extract visual tokens $z_0^{i}$ from its last $n$ frames as contextual conditioning for generating the next clip.
For the current clip $(i+1)$, we add noise to its visual tokens following the standard diffusion process to obtain $z_t^{i+1}$.
These tokens are concatenated along the sequence dimension as $z_t=[z_0^{i}; z_t^{i+1}]\in\mathbb{R}^{q \times c}$, where $q$ represents the total token count after concatenation.
We introduce a binary mask $m\in\mathbb{R}^{q \times 1}$ (1 for conditioning tokens, 0 for tokens being generated) and concatenate it with $z_t$ along the channel dimension to form $z_t=[z_t;m]\in \mathbb{R}^{q\times(c+1)}$.
The model from \cref{subsec:model} takes this combined feature with corresponding Plücker embeddings to predict the added noise, computing the loss from \cref{equ:diffusion_obj_control} only over tokens from the generated clip.
This process is shown in \cref{fig: modelarch} (b).
During inference, given a new camera trajectory, we select predefined frames from the previously generated clip as conditioning, enabling users to explore generated scenes through sequential camera trajectories while maintaining visual consistency between consecutive clips.

In sequential video generation, we use the first frame of the initial trajectory as the reference for calculating relative poses across all generated clips.
This unified coordinate system ensures geometric consistency throughout the sequence and prevents pose error accumulation between clips.

\begin{table}[t]
\centering \scriptsize
\setlength{\tabcolsep}{4pt}
\begin{tabular}{l c c c}
\toprule
Data Pipeline & TransErr$\downarrow$ & RotErr $\downarrow$ & Sample time (s) $\downarrow$ \\
\midrule
Before distillation & \textbf{0.1892} & \textbf{1.66} & 13.83 \\
Progressive dist.~\cite{progressivedistill} & 0.2001 & 1.90 & 2.61\\
APT~\cite{apt} & 0.2500 & 2.56 & \textbf{0.59}\\
\bottomrule
\end{tabular}
\caption{Model comparison before and after the distillations. 
The inference time is tested when generating a 4 second 12fps video with 4 H800 GPUs.} 
\vspace{-1.6em}
\label{tab:distill_comp}
\end{table}
\noindent\textbf{Model Distillation for Speedup.} 
To accelerate the inference speed and improve user experience, we implement a two-phase distillation approach.
First, we employed progressive distillation~\cite{progressivedistill} to reduce the required neural function evaluations~(NFEs) from 96 to 16 while maintaining visual quality.
The original 96 NFEs consisted of 32 for unconditional generation, 32 for text cfg generation, and 32 for camera cfg generation.
As shown in~\cref{tab:distill_comp}, the distilled model does not exist significant degradation in terms of camera control accuracy.
When generating a 4 second videos in 12fps with 4 H800 GPUs, the sample time is decreased significantly, from 13.83 second to 2.61 second.
This sample time contains the DiT model inference time and the VAE decode time.

To further accelerate, we apply the recent proposed distillation method APT~\cite{apt} for one-step generation.
~\cref{tab:distill_comp} presents the quality and speedup after distillation. 
Obviously, APT~\cite{apt} offers a significant speedup yet results in the degradation of conditional generation.
Considering the original APT~\cite{apt} that leverages more than one thousand GPUs, more computational resources and larger batch size could further improve the synthesis quality which we leave for the future.

\section{Experiments} \label{sec:exp}

\begin{table*}[t]
\centering \small
\setlength{\tabcolsep}{6pt}
\begin{tabular}{l c c c c c c}
\toprule
Model & FVD$_\downarrow$ & Motion strength$_\uparrow$ & TransErr$_\downarrow$ & RotErr$_\downarrow$  & Geometric consistency$_\uparrow$  & Appearance consistency$_\uparrow$ \\
\midrule
MotionCtrl~\cite{motionctrl} & 221.23 & 102.21 & 0.3221 & 2.78 & 57.87 & 0.7431 \\
CameraCtrl~\cite{cameractrl} & 199.53 & 133.37 & 0.2812 & 2.81 & 52.12 & 0.7784 \\
\method & \textbf{73.11} & \textbf{698.51} & \textbf{0.1527} & \textbf{1.58} & \textbf{88.70} & \textbf{0.8893} \\
\midrule
AC3D~\cite{ac3d} & 987.34 & 162.21 & 0.2976 & 2.98 & 69.20 & N/A\\
\method & \textbf{641.23} & \textbf{574.21} & \textbf{0.1892} & \textbf{1.66} & \textbf{85.00} & N/A \\
\bottomrule
\end{tabular}
\caption{\textbf{Quantitative Comparisons.} We compare against MotionCtrl~\cite{motionctrl} and CameraCtrl~\cite{cameractrl} in image-to-video setting, the AC3D~\cite{ac3d} in the text-to-video setting. Since open-sourcing AC3D only supports text-to-video generation, appearance consistency between given image and generated videos is not available.
}
\vspace{-1em}
\label{tab:quantitative_comparison}
\end{table*}

This section presents a comprehensive evaluation of \method, comparing it with existing approaches and validating its design choices.
This section is thus organized as follows.
Implementation details are provided in \cref{subsec:implement}.
The evaluation metrics and dataset specifications are described in \cref{subsec:metric} and \cref{subsec:eval_dataset}, respectively.
In \cref{subsec:compare}, we compare \method with other methods~\cite{cameractrl, motionctrl, ac3d}.
\cref{subsec:ablation} presents detailed ablation studies.
Finally, we provide some visualization results of \method in~\cref{subsec:visualizations}.

\subsection{Implementation Details} \label{subsec:implement}
Our model is based on an internal transformer-based text-to-video diffusion model, with approximately 3B parameters.
As a latent diffusion model, it employs a temporal causal VAE tokenizer similar to MAGViT2~\cite{magvit2}, with downsampling rate 4 for temporal and 8 for spatial.
We sample the camera poses every 4 frames, resulting in the same number of camera poses to the visual features.
During training, we kept all base video diffusion model parameters unfrozen, allowing joint optimization of all parameters.
We trained the model in two phases.
First phase is for the single clip \method~(\cref{subsec:model}) at a resolution of 192 $\times$ 320 for 100,000 steps with a batch size of 640, using video clips ranging from 2 to 10 seconds in duration.
The data composition maintains a 4:1 ratio between camera-labeled and unlabeled data.
In the second phase, we finetuned the model at a higher resolution of 384 $\times$ 640 while simultaneously training the video extension~(\cref{subsec:v2v_ext}).
This phase ran for 50,000 steps with a batch size of 512.
The number of condition frames from the previous clip ranges from a minimum of 5 frames to a maximum of 50\% of the total frames.
Both training stages utilize the AdamW optimizer.
The learning rate was initially set to $1 \times 10^{-4}$, with a warm-up period from $5 \times 10^{-5}$ over 500 steps.
, weight decay of 0.01, and betas of 0.9 and 0.95.
The learning rate was finally decayed to $1 \times 10^{-5}$ using the cosine learning rate scheduler.
We use 64 H100 GPUs for the first stage and 128 H100 GPUs for the second stage.
During the inference, we adopted the Euler sampler with 32 steps and a shift of 12~\cite{flux2024}.
We set the CFG scales to 7.5 and 8.0 for text and camera, respectively.

\subsection{Evaluation Metric}  \label{subsec:metric}
We utilize six metrics to comprehensively evaluate different aspects of baselines and our method, more details in appendix.
1) \textit{Visual Quality}: We adopt  Fréchet Video Distance~(FVD)~\cite{fvd} to measure the overall quality of the generated videos.
2) \textit{Video Dynamic Fidelity}: We propose motion strength to assess the dynamic degree of generated videos. 
This quantitative measure calculates the average motion magnitude of foreground objects across video frames using RAFT-extracted dense optical flow fields.
To isolate object motion from camera movement, we apply TMO-generated segmentation masks to the flow fields, computing motion for each lefted pixel as $\sqrt{u^2+^2}$ and converting from radians to degrees.
The final motion strength represents the average flow magnitude across all foreground pixels in all frames.
3) \textit{Camera Control Accuracy}: Following CameraCtrl~\cite{cameractrl}, we use TransErr and RotErr to measure the alignment between the condition camera poses and estimated camera poses from generated frames.
We extract motion patterns from generated videos using TMO~\cite{tmo} and estimate camera parameters with VGGSfM~\cite{wang2024vggsfm}. 
To address the inherent scale ambiguity in SfM, we align the estimated camera trajectory to ground truth using ATE~\cite{ate} by centering both trajectories, finding the optimal scale factor, computing rotation via SVD, and determining alignment translation.
After alignment, we calculate TransErr as the average Euclidean distance between corresponding camera positions and RotErr as the average angular difference between corresponding camera orientations.
4) \textit{Geometry Consistency}: We apply the VGGSfM~\cite{wang2024vggsfm} on the generated videos, and calculate the successful ratio of VGGSfM to estimate camera parameters.
It indicates the quality of 3D geometry consistency of a generated scene.
5) \textit{Scene Appearance Coherence}: Exploring a scene requires the model to generate sequential video clips for the same scene given sequential camera trajectories.
To evaluate visual consistency between these clips, we first extract features using a pretrained~\cite{clip} visual encoder for each frame in a video clip, then average these features to obtain the video feature for each video clip.
After that, we compute the cosine similarity across different video features, and term this metric as appearance consistency.
\subsection{Evaluation Dataset} \label{subsec:eval_dataset}
Our evaluation dataset consists of 800 video clips from two sources: 240 videos sampled from the RealEstate10K~\cite{realestate10k} test set, and 560 videos from our processed real-world dynamic videos with camera annotations.
We sampled videos across different camera trajectory categories analyzed in~\cref{subsec:data}. 

\begin{figure*}[t]
	\centering
	\includegraphics[width=\linewidth]{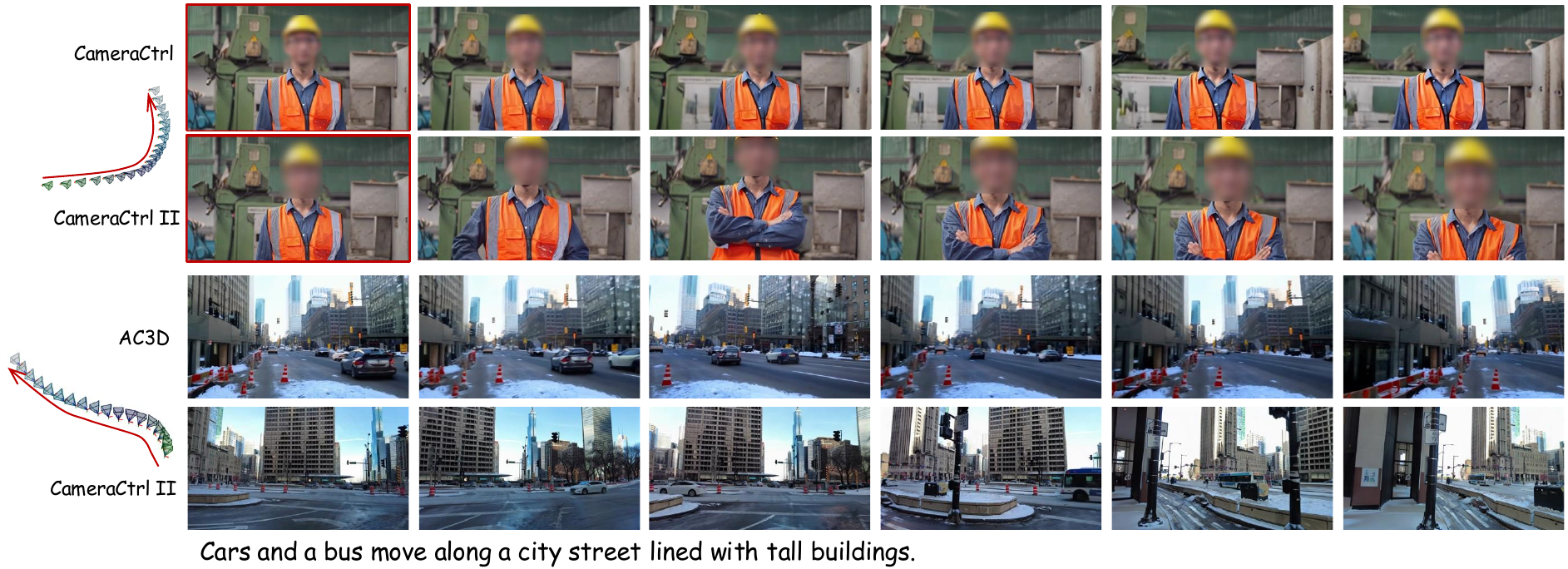}
	\caption{
        \textbf{Qualitative results.}
        The camera trajectories are shown in the left.
        The first two rows share the same camera trajectory, and the second camera trajectory is for the last two rows.
        We compare \method with CameraCtrl~\cite{cameractrl} in the I2V setting (first two rows), with the first image being the condition image.
        We also compare it with AC3D~\cite{ac3d} in the T2V setting in the last two rows.
        In both case, \method strictly follow each part of the camera trajectory and has better video dynamic.
        While CameraCtrl ignores the upward camera movements, AC3D ignores the forward camera moving at the end of the trajectory.
        }
	\label{fig: qualitative com}
\end{figure*}

\subsection{Comparisons with other methods} \label{subsec:compare}
\noindent\textbf{Quantitative comparison}. To evaluate the effectiveness of \method, we compare it with two representative methods, MotionCtrl~\cite{motionctrl} and CameraCtrl~\cite{cameractrl}, in the I2V setting.
Since these two methods cannot directly generate new video clips based on previously generated ones, we use the last frame of the previous video clip as the condition image to generate the next clip.
In addition, benefiting from our minimal modifications to the base model architecture and joint training strategy, our method can also be applied to camera-controlled T2V generation.
Thus, we compare \method with AC3D~\cite{ac3d} in the camera-controlled T2V task.
Due to the different number of camera parameters supported by these methods, we temporally downsample the camera parameters for them as the camera input.
As shown in the \cref{tab:quantitative_comparison}, \method significantly outperforms previous method across all metrics in both setting.
In the I2V setting, our method achieves better FVD, and higher Motion strength compared to MotionCtrl and CameraCtrl.
The camera control accuracy and geometric consistency are also improved, as indicated by lower TransErr, RotErr and higher Geometric consistency.
Similar improvements can be observed in the T2V setting when compared with AC3D.

\noindent\textbf{Qualitative Comparison.} 
We also provide qualitative comparisons in \cref{fig: qualitative com}.
As illustrated by the first two rows~(I2V setting), \method more accurately follows the input camera trajectories, while CameraCtrl~\cite{cameractrl} ignores the upward camera movements.
Besides, \method is able to generate more dynamic videos, while CameraCtrl tends to generate static ones.
The third and fourth rows compare \method with AC3D~\cite{ac3d} in the T2V setting.
\method effectively combines camera control with object motion, successfully generating dynamic elements such as moving vehicles.
In contrast, AC3D ignores the forward camera moving and does not strictly follow the text prompt, failed to generate a bus.

\begin{table}[t]
\centering 
\scriptsize
\setlength{\tabcolsep}{2pt}
\begin{tabular}{l c c c c }
\toprule
Model& Motion strength$_\uparrow$ & TransErr$_\downarrow$ & RotErr$_\downarrow$  & Geometric Consistency$_\uparrow$\\
\midrule
w/o Dyn. Vid & 129.40 & 0.2069 & 2.02 & 78.50\\
w/o Scale Calib.  & 301.68 & 0.2121 & 2.14 &82.10 \\
w/o Dist. Balance & \textbf{309.24} &0.2834 &4.56 & 85.96\\
\midrule
Full Pipeline & 306.99 & \textbf{0.1830} & \textbf{1.74} & \textbf{86.50}\\
\bottomrule
\end{tabular}
\caption{\textbf{Ablation study} on dataset curation pipeline.} 
\label{tab:ablation_data}
\end{table}

\subsection{Ablation Study} \label{subsec:ablation}
The design of \method consists of three key components: dataset construction, injecting the camera control into the pretrained video diffusion model, and a multi-clip video extension method.
In this section, we conduct extensive ablation studies to validate each component.
All models are trained at a resolution of 192 × 384 for 50,000 steps of single-clip training followed by 30,000 steps of multi-clip video extension training with the same resolution.

\noindent\textbf{Effectiveness of Each Component of Data Construction Pipeline.}
First, we investigate the necessity of incorporating dynamic videos by training only with the static data~(\cref{tab:ablation_data} w/o Dyn. Vid), RealEstate10K~\cite{realestate10k}.
The model shows degraded performance in terms of Motion strength~(129.40 vs 306.99) and camera control ability.
This  demonstrates that using dynamic videos with camera pose annotation during the training of camera-controlled video diffusion model is crucial for achieving high quality and high dynamic generation while maintaining camera control.

We then examine the importance of scale calibration by removing this step~(\cref{tab:ablation_data} w/o Scale Calib.).
The results show that without this step, the model exhibits higher camera  control errors~(TransErr 0.2121 vs 0.1830, RotErr 2.14 vs 1.74), and the lower Geometric consistency.
This validates our hypothesis that normalizing scene scales to the same metric space helps the model learn more consistent geometric relationships, contributing to more accurate camera control and easier scene reconstruction.

After that, we analyze the effect of distribution balancing in camera trajectory types~(\cref{tab:ablation_data} w/o Dist. Balance). 
Without this step, the model shows notable degradation in camera control accuracy and geometric consistency.
This confirms that balancing the extreme long-tailed distribution of camera trajectory distributions of real-world videos is essential for achieving robust camera control and geometric consistency across diverse camera movement patterns.

\begin{table}[t]
\centering \scriptsize
\setlength{\tabcolsep}{2pt}
\begin{tabular}{l c c c c}
\toprule
Model& Motion strength$_\uparrow$ & TransErr$_\downarrow$ & RotErr$_\downarrow$  & Geometric consistency$_\uparrow$ \\
\midrule
Complex Encoder & 301.23 & \textbf{0.1826} & 1.88 & 84.00 \\
Multilayer Inj. & 247.23 & 0.1865 & 1.78 & 85.00 \\
w/o Joint Training  & 279.82 & 0.2098 & 1.97 & 81.92 \\
\midrule
\method & \textbf{306.99} & 0.1830 &\textbf{1.74} & \textbf{86.50} \\
\bottomrule
\end{tabular}
\caption{\textbf{Ablation study} on the effectiveness of our model architecture and training strategy for single-clip model.}
\label{tab:ablation_model_arch}
\end{table}

\noindent\textbf{Effectiveness of Model Design and Training Strategy.}
Next, we conduct ablation studies on our design choice for the camera pose injection module and the training strategy.
First, we evaluate the effectiveness of using a single patchify layer to extract camera features from the Plücker embedding.
For comparison, we implement a model variant with a more sophisticated encoder similar to CameraCtrl, and the extracted camera features are used at the beginning of the DiT model.
As shown in \cref{tab:ablation_model_arch} (Complex Encoder),  while this more complex architecture achieves comparable performance in terms of TransErr, our simple patchify layer design yields better results across other metrics.
This finding indicates that a simple feature extraction layer is sufficient for converting camera representations into effective guidance signals for the generation process.

We then investigate the impact of camera condition injection places.
While injecting camera features at every DiT layers~(\cref{tab:ablation_model_arch} Multilayer Inj.) achieves comparable camera control accuracy, it significantly reduce the Motion strength.
This result supports our argument that camera control information should only guide the overall video generation.
Adding camera features to the deeper layers, where the model processes local details, can restrict the model's capability to generate dynamic videos.
Thus, adding the camera representation at the initial layer of DiT model is sufficient.

Next, we study the effectiveness of joint training the model with additional video data without camera annotations~(\cref{tab:ablation_model_arch} w/o Joint Training).
Results show that removing this joint training leads to reduced dynamics.
This is because additional video data exposes our model to more diverse visual domains and object motion types that are not covered in the RealCam dataset.
Additionally, joint training helps improve camera control performance by enabling the camera-wise classifier-free guidance, as demonstrated by TransErr, RotErr, and Geometric consistency.

\begin{table}[t]
\centering \scriptsize
\setlength{\tabcolsep}{4pt}
\begin{tabular}{l c c c c}
\toprule
Model & FVD$_\downarrow$& TransErr$_\downarrow$ & RotErr$_\downarrow$  & Appearance consistency$_\uparrow$\\
\midrule
Different Ref.  & 118.32 & 0.1963 & 1.94 & 0.8032 \\
Noised Condition & 136.78 & 0.1847 & 1.85 & 0.7843 \\
Noised Condition$^{*}$ & 140.98 & 0.1901 & 1.88 & 0.7982 \\
\midrule
\method & \textbf{112.46} & \textbf{0.1830} & \textbf{1.74} & \textbf{0.8654} \\
\bottomrule
\end{tabular}
\caption{\textbf{Ablation study} on key design choices in extending the single-clip model to enable scene exploration.}
\label{tab:ablation_extension}
\end{table}

\begin{figure*}[t]
	\centering
	\includegraphics[width=\linewidth]{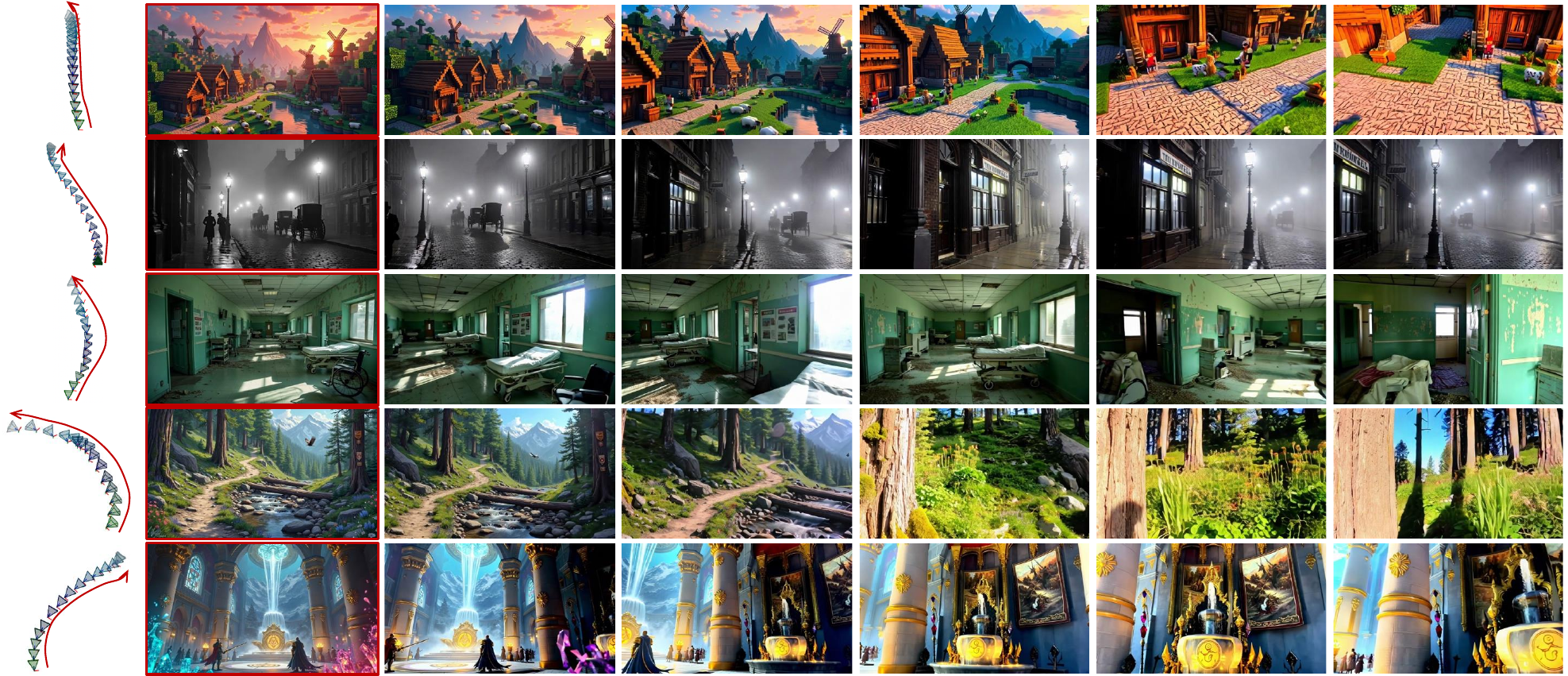}
	\caption{
        \textbf{Visualization results of \method across diverse scenes.}
        Our model demonstrates effective camera control in various visual environments, including Minecraft-style game scenes (top row), black and white foggy London streets (second row), abandoned hospital interiors (third row), fantasy forest hiking trails (fourth row), and animated palace scenes (bottom row).
        The results are generated using the I2V setting, with the first image as the condition image.
        The camera trajectories are shown on the left of each row.
        }
	\label{fig: visualization}
\end{figure*}

\begin{figure*}[t]
    \centering
    \includegraphics[width=\linewidth]{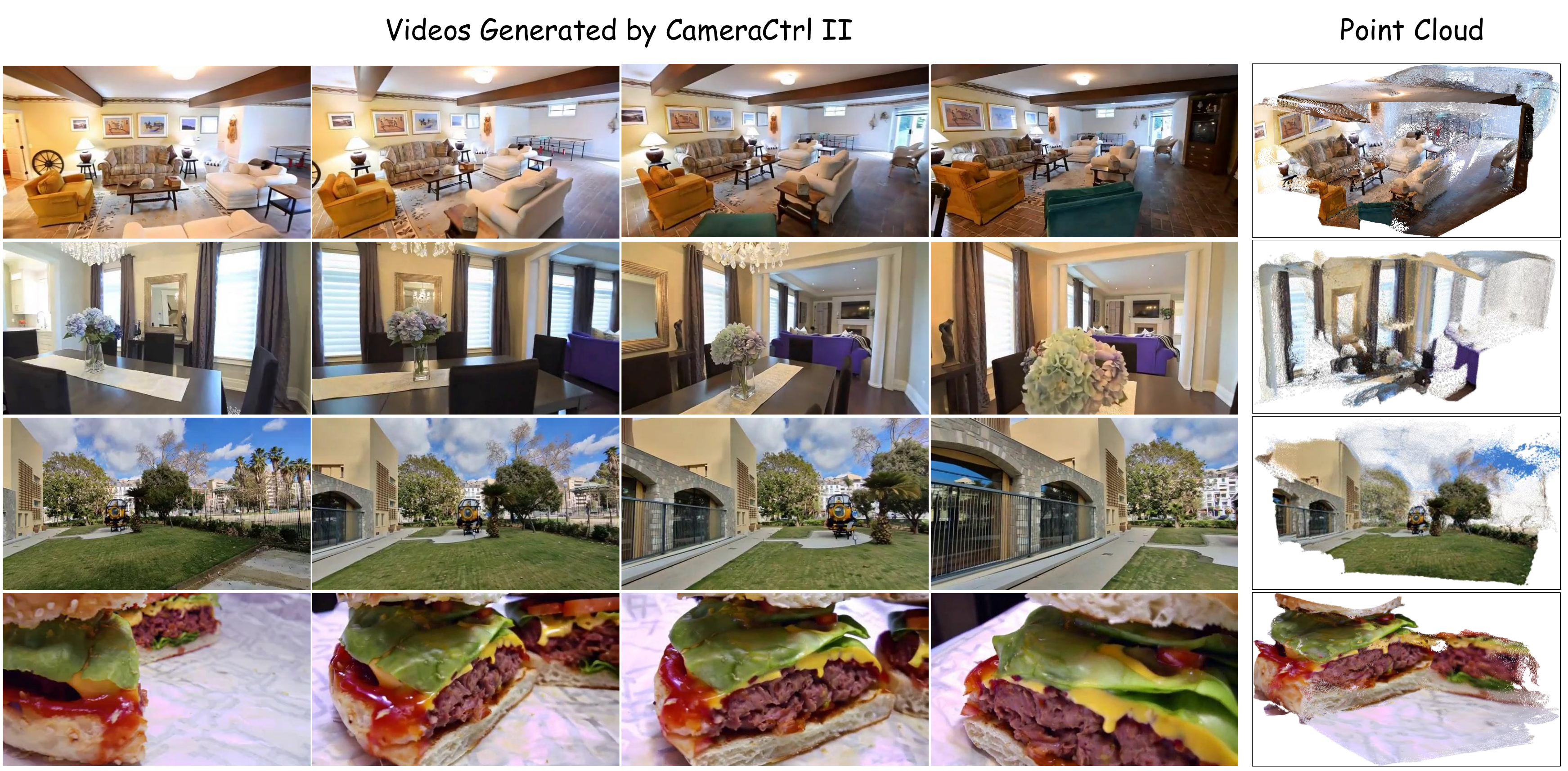}
    \caption{\textbf{3D reconstruction} on generated scenes by \method. 
With the generated video frames, we use the FLARE~\cite{zhang2025flare} to estimate the point clouds of the scenes.}
    \label{fig:3drecon}
\end{figure*}

\noindent\textbf{Key Design Choices for Video Extension.}
Finally, we investigate two important design choices for video extension.
First, we study different strategy for defining reference frames for calculating the relative camera pose.
One approach uses each clip's first frame as a local reference frame within that clip~(\cref{tab:ablation_extension} Different Ref.).
Our method uses the first frame of the first clip as a global reference frame to compute the relative camera pose for all camera trajectories. 
Results show that using a global reference achieves better camera control accuracy and Appearance consistency.
This is because that a shared reference frame helps maintain consistent geometric relationships across clips and camera trajectory conditions, making it easier for the model to learn smooth transition between clips.

We compare our clip-wise extension approach with an alternative strategy.
In this alternative model (\cref{tab:ablation_extension} Noised Condition), noise is added to all clips during training, and the loss is computed across both conditioning and target clips.
However, during inference, only clean conditioning clips are used, creating a discrepancy between training and inference settings.
This mismatch leads to degraded performance in both FVD and Appearance consistency metrics.
Even when attempting to bridge this gap by adding little noise to conditioning frames in inference, the performance remains suboptimal (\cref{tab:ablation_extension} Noised Condition').
In contrast, our teacher maintains consistent noise-free condition clips in both training and inference.
\subsection{Visualization Results} \label{subsec:visualizations}
\noindent \textbf{Different scenario scenes exploration.}
We first provide visualizations of \method for different scene scenarios to showcase its generalization performance in camera control.
As shown in the \cref{fig: visualization}, our model can be applied to various scenes (such as Minecraft-like game scenes, black and white 19th century foggy London streets, indoor abandoned hospital, outdoor hiking in a fantasy world, and anime-style palace scenes).
Besides, \method can effectively controlling camera movements (camera panning left and right, complete turns, etc.) and maintaining appropriate dynamic effects.

\noindent \textbf{3D reconstruction of generated scenes.} Our method generates high-quality dynamic videos with conditional camera poses, effectively transforming video generative models into view synthesizers.
The strong 3D consistency of these generated videos enables high-quality 3D reconstruction. 
Specifically, we use FLARE~\cite{zhang2025flare} to infer detailed 3D point clouds from frames extracted from our generated videos.
As shown in \cref{fig:3drecon}, our approach produces videos that can be reconstructed into high-quality point clouds, demonstrating the superior 3D consistency achieved by our models.

\begin{figure*}[t]
	\centering
	\includegraphics[width=\linewidth]{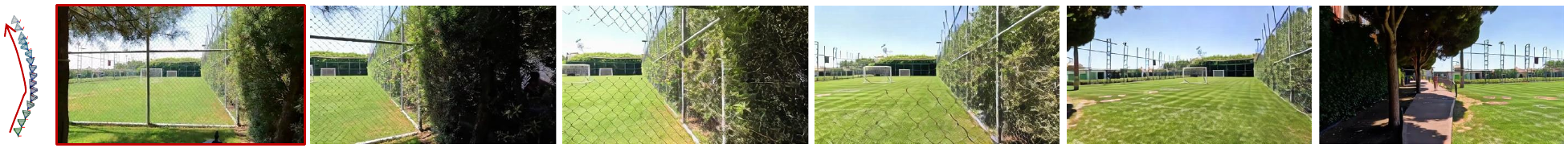}
	\caption{\textbf{Failure case visualization}.
    A fence is on the intended camera trajectory.
    \method strictly follows the trajectory and generate a video where the structure of the fence is damaged, which is anti-reality.
    }
	\label{fig:failure_case}
\end{figure*}

\section{Discussion} \label{sec:conclusion}
In this paper, we introduce \method, a framework that enables users to explore generated dynamic scenes through precise camera control.
We first construct \dataset, a dataset composed of dynamic videos with camera pose annotations.
Then, we design a lightweight camera parameter injector that integrates camera conditions at the initial layers of DiT, along with a corresponding joint training strategy to preserve the pre-trained model's ability to generate dynamic scenes.
Besides, we develop a clip-level extension method that allows the model to generate new video clips conditioned on both previously generated content and new camera trajectories.
Experimental results demonstrate our method's effectiveness in generating camera-controlled dynamic videos while maintaining high quality and temporal consistency across sequential video clips.

\noindent\textbf{Limitation and Future Work.}
Our current approach has several limitations for future investigation. 
First, \method occasionally struggles to resolve conflicts between camera movement and scene geometry, sometimes resulting in physically implausible camera paths that intersect with scene structures. 
We provide a failure case in~\cref{fig:failure_case}.
In this example, we provide a forward camera trajectory with left and right camera view change.
There is a fence blocks the intended path.
An ideal physically-aware model would recognize this constraint and stop the camera movement at the fence. 
However, our model generates a physically implausible result where the fence structure deteriorates as the camera passes through it.
Additionally, while our method achieves accurate camera control, the overall geometric consistency of generated scenes could be further improved, especially when dealing with complex camera trajectories.

\noindent \textbf{Ethics Statement.} 
Our camera trajectory-based video generation model enables dynamic scene generation and exploration, but we acknowledge potential ethical concerns.
While this technology has beneficial applications in education, virtual tourism, and creative industries, it could potentially be misused to create misleading content.
Our model's outputs may reflect biases present in the training data despite our efforts to use diverse training sets.
Users should obtain proper consent when using personal images as input and respect copyright, privacy, and cultural sensitivities when utilizing our system.

\section{Acknowledgement}\label{sec:ack}

We would like to express our sincere gratitude to Jianyuan Wang in the University of Oxford for his valuable insights and assistance regarding the usage of VGGSfM in our data processing pipeline.
His expertise significantly contributed to the development of our camera pose estimation approach and enhanced the overall quality of our work.

\clearpage

{
    \small
    \bibliographystyle{ieeenat_fullname}
    \bibliography{main}

\begin{thebibliography}{62}
\providecommand{\natexlab}[1]{#1}
\providecommand{\url}[1]{\texttt{#1}}
\expandafter\ifx\csname urlstyle\endcsname\relax
  \providecommand{\doi}[1]{doi: #1}\else
  \providecommand{\doi}{doi: \begingroup \urlstyle{rm}\Url}\fi

\bibitem[noa()]{noauthor_cosmos_nodate}
Cosmos {World} {Foundation} {Model} {Platform} for {Physical} {AI}.

\bibitem[Bahmani et~al.(2024{\natexlab{a}})Bahmani, Skorokhodov, Qian, Siarohin, Menapace, Tagliasacchi, Lindell, and Tulyakov]{ac3d}
Sherwin Bahmani, Ivan Skorokhodov, Guocheng Qian, Aliaksandr Siarohin, Willi Menapace, Andrea Tagliasacchi, David~B Lindell, and Sergey Tulyakov.
\newblock Ac3d: Analyzing and improving 3d camera control in video diffusion transformers.
\newblock \emph{arXiv preprint arXiv:2411.18673}, 2024{\natexlab{a}}.

\bibitem[Bahmani et~al.(2024{\natexlab{b}})Bahmani, Skorokhodov, Siarohin, Menapace, Qian, Vasilkovsky, Lee, Wang, Zou, Tagliasacchi, Lindell, and Tulyakov]{bahmani2024vd3d}
Sherwin Bahmani, Ivan Skorokhodov, Aliaksandr Siarohin, Willi Menapace, Guocheng Qian, Michael Vasilkovsky, Hsin-Ying Lee, Chaoyang Wang, Jiaxu Zou, Andrea Tagliasacchi, David~B. Lindell, and Sergey Tulyakov.
\newblock Vd3d: Taming large video diffusion transformers for 3d camera control.
\newblock \emph{arXiv preprint arXiv:2407.12781}, 2024{\natexlab{b}}.

\bibitem[Bai et~al.(2024)Bai, Xia, Wang, Yuan, Fu, Liu, Hu, Wan, and Zhang]{bai2024syncammaster}
Jianhong Bai, Menghan Xia, Xintao Wang, Ziyang Yuan, Xiao Fu, Zuozhu Liu, Haoji Hu, Pengfei Wan, and Di Zhang.
\newblock Syncammaster: Synchronizing multi-camera video generation from diverse viewpoints.
\newblock \emph{arXiv preprint arXiv:2412.07760}, 2024.

\bibitem[Bain et~al.(2021)Bain, Nagrani, Varol, and Zisserman]{webvid10m}
Max Bain, Arsha Nagrani, G{\"u}l Varol, and Andrew Zisserman.
\newblock Frozen in time: A joint video and image encoder for end-to-end retrieval.
\newblock In \emph{Proceedings of the IEEE/CVF international conference on computer vision}, pages 1728--1738, 2021.

\bibitem[Blattmann et~al.(2023{\natexlab{a}})Blattmann, Dockhorn, Kulal, Mendelevitch, Kilian, Lorenz, Levi, English, Voleti, Letts, et~al.]{svd}
Andreas Blattmann, Tim Dockhorn, Sumith Kulal, Daniel Mendelevitch, Maciej Kilian, Dominik Lorenz, Yam Levi, Zion English, Vikram Voleti, Adam Letts, et~al.
\newblock Stable video diffusion: Scaling latent video diffusion models to large datasets.
\newblock \emph{arXiv preprint arXiv:2311.15127}, 2023{\natexlab{a}}.

\bibitem[Blattmann et~al.(2023{\natexlab{b}})Blattmann, Rombach, Ling, Dockhorn, Kim, Fidler, and Kreis]{alignyourlatents}
Andreas Blattmann, Robin Rombach, Huan Ling, Tim Dockhorn, Seung~Wook Kim, Sanja Fidler, and Karsten Kreis.
\newblock Align your latents: High-resolution video synthesis with latent diffusion models.
\newblock In \emph{Proceedings of the IEEE/CVF conference on computer vision and pattern recognition}, pages 22563--22575, 2023{\natexlab{b}}.

\bibitem[Bochkovskii et~al.(2024)Bochkovskii, Delaunoy, Germain, Santos, Zhou, Richter, and Koltun]{depthpro}
Aleksei Bochkovskii, Ama{\"e}l Delaunoy, Hugo Germain, Marcel Santos, Yichao Zhou, Stephan~R Richter, and Vladlen Koltun.
\newblock Depth pro: Sharp monocular metric depth in less than a second.
\newblock \emph{arXiv preprint arXiv:2410.02073}, 2024.

\bibitem[Brooks et~al.(2024)Brooks, Peebles, Holmes, DePue, Guo, Jing, Schnurr, Taylor, Luhman, Luhman, Ng, Wang, and Ramesh]{sora}
Tim Brooks, Bill Peebles, Connor Holmes, Will DePue, Yufei Guo, Li Jing, David Schnurr, Joe Taylor, Troy Luhman, Eric Luhman, Clarence Ng, Ricky Wang, and Aditya Ramesh.
\newblock Video generation models as world simulators.
\newblock 2024.

\bibitem[Chen et~al.(2024)Chen, Siarohin, Menapace, Deyneka, Chao, Jeon, Fang, Lee, Ren, Yang, et~al.]{panda70m}
Tsai-Shien Chen, Aliaksandr Siarohin, Willi Menapace, Ekaterina Deyneka, Hsiang-wei Chao, Byung~Eun Jeon, Yuwei Fang, Hsin-Ying Lee, Jian Ren, Ming-Hsuan Yang, et~al.
\newblock Panda-70m: Captioning 70m videos with multiple cross-modality teachers.
\newblock In \emph{Proceedings of the IEEE/CVF Conference on Computer Vision and Pattern Recognition}, pages 13320--13331, 2024.

\bibitem[Cho et~al.(2023)Cho, Lee, Lee, Park, Kim, and Lee]{tmo}
Suhwan Cho, Minhyeok Lee, Seunghoon Lee, Chaewon Park, Donghyeong Kim, and Sangyoun Lee.
\newblock Treating motion as option to reduce motion dependency in unsupervised video object segmentation.
\newblock In \emph{Proceedings of the IEEE/CVF winter conference on applications of computer vision}, pages 5140--5149, 2023.

\bibitem[Deitke et~al.(2023)Deitke, Liu, Wallingford, Ngo, Michel, Kusupati, Fan, Laforte, Voleti, Gadre, et~al.]{objaverse}
Matt Deitke, Ruoshi Liu, Matthew Wallingford, Huong Ngo, Oscar Michel, Aditya Kusupati, Alan Fan, Christian Laforte, Vikram Voleti, Samir~Yitzhak Gadre, et~al.
\newblock Objaverse-xl: A universe of 10m+ 3d objects.
\newblock \emph{Advances in Neural Information Processing Systems}, 36:\penalty0 35799--35813, 2023.

\bibitem[Esser et~al.(2024)Esser, Kulal, Blattmann, Entezari, M{\"u}ller, Saini, Levi, Lorenz, Sauer, Boesel, et~al.]{sd3}
Patrick Esser, Sumith Kulal, Andreas Blattmann, Rahim Entezari, Jonas M{\"u}ller, Harry Saini, Yam Levi, Dominik Lorenz, Axel Sauer, Frederic Boesel, et~al.
\newblock Scaling rectified flow transformers for high-resolution image synthesis.
\newblock In \emph{Forty-first international conference on machine learning}, 2024.

\bibitem[Feng et~al.(2024)Feng, Zhang, Yang, Xiao, Shu, Liu, Zheng, Huang, Liu, and Zhang]{thematrix}
Ruili Feng, Han Zhang, Zhantao Yang, Jie Xiao, Zhilei Shu, Zhiheng Liu, Andy Zheng, Yukun Huang, Yu Liu, and Hongyang Zhang.
\newblock The matrix: Infinite-horizon world generation with real-time moving control.
\newblock \emph{arXiv preprint arXiv:2412.03568}, 2024.

\bibitem[Feng et~al.(2025)Feng, Liu, Tu, Qi, Sun, Ma, Zhao, Zhou, and He]{i2vcontrolcamera}
Wanquan Feng, Jiawei Liu, Pengqi Tu, Tianhao Qi, Mingzhen Sun, Tianxiang Ma, Songtao Zhao, Siyu Zhou, and Qian He.
\newblock I2vcontrol-camera: Precise video camera control with adjustable motion strength.
\newblock 2025.

\bibitem[Fu et~al.(2024)Fu, Liu, Wang, Peng, Xia, Shi, Yuan, Wan, Zhang, and Lin]{fu20243dtrajmaster}
Xiao Fu, Xian Liu, Xintao Wang, Sida Peng, Menghan Xia, Xiaoyu Shi, Ziyang Yuan, Pengfei Wan, Di Zhang, and Dahua Lin.
\newblock 3dtrajmaster: Mastering 3d trajectory for multi-entity motion in video generation.
\newblock \emph{arXiv preprint arXiv:2412.07759}, 2024.

\bibitem[Guo et~al.(2023)Guo, Yang, Rao, Liang, Wang, Qiao, Agrawala, Lin, and Dai]{guo2023animatediff}
Yuwei Guo, Ceyuan Yang, Anyi Rao, Zhengyang Liang, Yaohui Wang, Yu Qiao, Maneesh Agrawala, Dahua Lin, and Bo Dai.
\newblock Animatediff: Animate your personalized text-to-image diffusion models without specific tuning.
\newblock \emph{arXiv preprint arXiv:2307.04725}, 2023.

\bibitem[Guo et~al.(2024)Guo, Yang, Rao, Agrawala, Lin, and Dai]{sparsectrl}
Yuwei Guo, Ceyuan Yang, Anyi Rao, Maneesh Agrawala, Dahua Lin, and Bo Dai.
\newblock Sparsectrl: Adding sparse controls to text-to-video diffusion models.
\newblock In \emph{European Conference on Computer Vision}, pages 330--348. Springer, 2024.

\bibitem[Gupta et~al.(2024)Gupta, Yu, Sohn, Gu, Hahn, Li, Essa, Jiang, and Lezama]{gupta2024photorealistic}
Agrim Gupta, Lijun Yu, Kihyuk Sohn, Xiuye Gu, Meera Hahn, Fei-Fei Li, Irfan Essa, Lu Jiang, and Jos{\'e} Lezama.
\newblock Photorealistic video generation with diffusion models.
\newblock In \emph{European Conference on Computer Vision}, pages 393--411. Springer, 2024.

\bibitem[Hartley and Zisserman(2003)]{mvg}
Richard Hartley and Andrew Zisserman.
\newblock \emph{Multiple view geometry in computer vision}.
\newblock Cambridge university press, 2003.

\bibitem[He et~al.(2024)He, Xu, Guo, Wetzstein, Dai, Li, and Yang]{cameractrl}
Hao He, Yinghao Xu, Yuwei Guo, Gordon Wetzstein, Bo Dai, Hongsheng Li, and Ceyuan Yang.
\newblock Cameractrl: Enabling camera control for text-to-video generation.
\newblock \emph{arXiv preprint arXiv:2404.02101}, 2024.

\bibitem[Ho and Salimans(2022)]{cfg}
Jonathan Ho and Tim Salimans.
\newblock Classifier-free diffusion guidance.
\newblock \emph{arXiv preprint arXiv:2207.12598}, 2022.

\bibitem[Ho et~al.(2022)Ho, Salimans, Gritsenko, Chan, Norouzi, and Fleet]{ho2022vdm}
Jonathan Ho, Tim Salimans, Alexey Gritsenko, William Chan, Mohammad Norouzi, and David~J Fleet.
\newblock Video diffusion models.
\newblock \emph{Advances in Neural Information Processing Systems}, 35:\penalty0 8633--8646, 2022.

\bibitem[Hong et~al.(2022)Hong, Ding, Zheng, Liu, and Tang]{cogvideo}
Wenyi Hong, Ming Ding, Wendi Zheng, Xinghan Liu, and Jie Tang.
\newblock Cogvideo: Large-scale pretraining for text-to-video generation via transformers.
\newblock \emph{arXiv preprint arXiv:2205.15868}, 2022.

\bibitem[Hou et~al.(2024)Hou, Wei, Zeng, and Chen]{camtrol}
Chen Hou, Guoqiang Wei, Yan Zeng, and Zhibo Chen.
\newblock Training-free camera control for video generation.
\newblock \emph{arXiv preprint arXiv:2406.10126}, 2024.

\bibitem[Huang et~al.(2024{\natexlab{a}})Huang, He, Yu, Zhang, Si, Jiang, Zhang, Wu, Jin, Chanpaisit, Wang, Chen, Wang, Lin, Qiao, and Liu]{vbench}
Ziqi Huang, Yinan He, Jiashuo Yu, Fan Zhang, Chenyang Si, Yuming Jiang, Yuanhan Zhang, Tianxing Wu, Qingyang Jin, Nattapol Chanpaisit, Yaohui Wang, Xinyuan Chen, Limin Wang, Dahua Lin, Yu Qiao, and Ziwei Liu.
\newblock {VBench}: Comprehensive benchmark suite for video generative models.
\newblock In \emph{Proceedings of the IEEE/CVF Conference on Computer Vision and Pattern Recognition}, 2024{\natexlab{a}}.

\bibitem[Huang et~al.(2024{\natexlab{b}})Huang, Zhang, Xu, He, Yu, Dong, Ma, Chanpaisit, Si, Jiang, Wang, Chen, Chen, Wang, Lin, Qiao, and Liu]{vbench++}
Ziqi Huang, Fan Zhang, Xiaojie Xu, Yinan He, Jiashuo Yu, Ziyue Dong, Qianli Ma, Nattapol Chanpaisit, Chenyang Si, Yuming Jiang, Yaohui Wang, Xinyuan Chen, Ying-Cong Chen, Limin Wang, Dahua Lin, Yu Qiao, and Ziwei Liu.
\newblock Vbench++: Comprehensive and versatile benchmark suite for video generative models.
\newblock \emph{arXiv preprint arXiv:2411.13503}, 2024{\natexlab{b}}.

\bibitem[Karras et~al.(2022)Karras, Aittala, Aila, and Laine]{edm}
Tero Karras, Miika Aittala, Timo Aila, and Samuli Laine.
\newblock Elucidating the design space of diffusion-based generative models.
\newblock \emph{Advances in neural information processing systems}, 35:\penalty0 26565--26577, 2022.

\bibitem[KuaiShou(2025)]{kling}
KuaiShou.
\newblock Kling, 2025.
\newblock Accessed: 2025-02-23.

\bibitem[Kuang et~al.(2025)Kuang, Cai, He, Xu, Li, Guibas, and Wetzstein]{kuang2025collaborative}
Zhengfei Kuang, Shengqu Cai, Hao He, Yinghao Xu, Hongsheng Li, Leonidas~J Guibas, and Gordon Wetzstein.
\newblock Collaborative video diffusion: Consistent multi-video generation with camera control.
\newblock \emph{Advances in Neural Information Processing Systems}, 37:\penalty0 16240--16271, 2025.

\bibitem[Labs(2024)]{flux2024}
Black~Forest Labs.
\newblock Flux.
\newblock \url{https://github.com/black-forest-labs/flux}, 2024.

\bibitem[Li et~al.(2025)Li, Zheng, Zhu, Mai, Zhang, Wonka, and Ghanem]{vividzoo}
Bing Li, Cheng Zheng, Wenxuan Zhu, Jinjie Mai, Biao Zhang, Peter Wonka, and Bernard Ghanem.
\newblock Vivid-zoo: Multi-view video generation with diffusion model.
\newblock \emph{Advances in Neural Information Processing Systems}, 37:\penalty0 62189--62222, 2025.

\bibitem[Lin et~al.(2024)Lin, Ge, Cheng, Li, Zhu, Wang, He, Ye, Yuan, Chen, et~al.]{lin2024open}
Bin Lin, Yunyang Ge, Xinhua Cheng, Zongjian Li, Bin Zhu, Shaodong Wang, Xianyi He, Yang Ye, Shenghai Yuan, Liuhan Chen, et~al.
\newblock Open-sora plan: Open-source large video generation model.
\newblock \emph{arXiv preprint arXiv:2412.00131}, 2024.

\bibitem[Lin et~al.(2025)Lin, Xia, Ren, Yang, Xiao, and Jiang]{apt}
Shanchuan Lin, Xin Xia, Yuxi Ren, Ceyuan Yang, Xuefeng Xiao, and Lu Jiang.
\newblock Diffusion adversarial post-training for one-step video generation.
\newblock \emph{arXiv preprint arXiv:2501.08316}, 2025.

\bibitem[Ling et~al.(2024)Ling, Sheng, Tu, Zhao, Xin, Wan, Yu, Guo, Yu, Lu, et~al.]{ling2024dl3dv}
Lu Ling, Yichen Sheng, Zhi Tu, Wentian Zhao, Cheng Xin, Kun Wan, Lantao Yu, Qianyu Guo, Zixun Yu, Yawen Lu, et~al.
\newblock Dl3dv-10k: A large-scale scene dataset for deep learning-based 3d vision.
\newblock In \emph{Proceedings of the IEEE/CVF Conference on Computer Vision and Pattern Recognition}, pages 22160--22169, 2024.

\bibitem[Liu et~al.(2021)Liu, Tucker, Jampani, Makadia, Snavely, and Kanazawa]{infinite_nature_2020}
Andrew Liu, Richard Tucker, Varun Jampani, Ameesh Makadia, Noah Snavely, and Angjoo Kanazawa.
\newblock Infinite nature: Perpetual view generation of natural scenes from a single image.
\newblock In \emph{Proceedings of the IEEE/CVF International Conference on Computer Vision (ICCV)}, 2021.

\bibitem[Liu et~al.(2022)Liu, Gong, and Liu]{recitfiedflow}
Xingchao Liu, Chengyue Gong, and Qiang Liu.
\newblock Flow straight and fast: Learning to generate and transfer data with rectified flow.
\newblock \emph{arXiv preprint arXiv:2209.03003}, 2022.

\bibitem[Ma et~al.(2025)Ma, Huang, Yan, Chen, Duan, Yin, Wan, Ming, Song, Chen, et~al.]{ma2025step}
Guoqing Ma, Haoyang Huang, Kun Yan, Liangyu Chen, Nan Duan, Shengming Yin, Changyi Wan, Ranchen Ming, Xiaoniu Song, Xing Chen, et~al.
\newblock Step-video-t2v technical report: The practice, challenges, and future of video foundation model.
\newblock \emph{arXiv preprint arXiv:2502.10248}, 2025.

\bibitem[Ma et~al.(2024)Ma, Wang, Jia, Chen, Liu, Li, Chen, and Qiao]{ma2024latte}
Xin Ma, Yaohui Wang, Gengyun Jia, Xinyuan Chen, Ziwei Liu, Yuan-Fang Li, Cunjian Chen, and Yu Qiao.
\newblock Latte: Latent diffusion transformer for video generation.
\newblock \emph{arXiv preprint arXiv:2401.03048}, 2024.

\bibitem[Menapace et~al.(2024)Menapace, Siarohin, Skorokhodov, Deyneka, Chen, Kag, Fang, Stoliar, Ricci, Ren, et~al.]{snapvideo}
Willi Menapace, Aliaksandr Siarohin, Ivan Skorokhodov, Ekaterina Deyneka, Tsai-Shien Chen, Anil Kag, Yuwei Fang, Aleksei Stoliar, Elisa Ricci, Jian Ren, et~al.
\newblock Snap video: Scaled spatiotemporal transformers for text-to-video synthesis.
\newblock In \emph{Proceedings of the IEEE/CVF Conference on Computer Vision and Pattern Recognition}, pages 7038--7048, 2024.

\bibitem[Parker-Holder et~al.(2024)Parker-Holder, Ball, Bruce, Dasagi, Holsheimer, Kaplanis, Moufarek, Scully, Shar, Shi, Spencer, Yung, Dennis, Kenjeyev, Long, Mnih, Chan, Gazeau, Li, Pardo, Wang, Zhang, Besse, Harley, Mitenkova, Wang, Clune, Hassabis, Hadsell, Bolton, Singh, and Rockt{\"a}schel]{parkerholder2024genie2}
Jack Parker-Holder, Philip Ball, Jake Bruce, Vibhavari Dasagi, Kristian Holsheimer, Christos Kaplanis, Alexandre Moufarek, Guy Scully, Jeremy Shar, Jimmy Shi, Stephen Spencer, Jessica Yung, Michael Dennis, Sultan Kenjeyev, Shangbang Long, Vlad Mnih, Harris Chan, Maxime Gazeau, Bonnie Li, Fabio Pardo, Luyu Wang, Lei Zhang, Frederic Besse, Tim Harley, Anna Mitenkova, Jane Wang, Jeff Clune, Demis Hassabis, Raia Hadsell, Adrian Bolton, Satinder Singh, and Tim Rockt{\"a}schel.
\newblock Genie 2: A large-scale foundation world model.
\newblock 2024.

\bibitem[Peebles and Xie(2023)]{dit}
William Peebles and Saining Xie.
\newblock Scalable diffusion models with transformers.
\newblock In \emph{Proceedings of the IEEE/CVF international conference on computer vision}, pages 4195--4205, 2023.

\bibitem[Radford et~al.(2021)Radford, Kim, Hallacy, Ramesh, Goh, Agarwal, Sastry, Askell, Mishkin, Clark, et~al.]{clip}
Alec Radford, Jong~Wook Kim, Chris Hallacy, Aditya Ramesh, Gabriel Goh, Sandhini Agarwal, Girish Sastry, Amanda Askell, Pamela Mishkin, Jack Clark, et~al.
\newblock Learning transferable visual models from natural language supervision.
\newblock In \emph{International conference on machine learning}, pages 8748--8763. PmLR, 2021.

\bibitem[Salimans and Ho(2022)]{progressivedistill}
Tim Salimans and Jonathan Ho.
\newblock Progressive distillation for fast sampling of diffusion models.
\newblock \emph{arXiv preprint arXiv:2202.00512}, 2022.

\bibitem[Shi et~al.(2024)Shi, Huang, Wang, Bian, Li, Zhang, Zhang, Cheung, See, Qin, et~al.]{xiaoyu_motioni2v}
Xiaoyu Shi, Zhaoyang Huang, Fu-Yun Wang, Weikang Bian, Dasong Li, Yi Zhang, Manyuan Zhang, Ka~Chun Cheung, Simon See, Hongwei Qin, et~al.
\newblock Motion-i2v: Consistent and controllable image-to-video generation with explicit motion modeling.
\newblock In \emph{ACM SIGGRAPH 2024 Conference Papers}, pages 1--11, 2024.

\bibitem[Singer et~al.(2022)Singer, Polyak, Hayes, Yin, An, Zhang, Hu, Yang, Ashual, Gafni, et~al.]{makeavideo}
Uriel Singer, Adam Polyak, Thomas Hayes, Xi Yin, Jie An, Songyang Zhang, Qiyuan Hu, Harry Yang, Oron Ashual, Oran Gafni, et~al.
\newblock Make-a-video: Text-to-video generation without text-video data.
\newblock \emph{arXiv preprint arXiv:2209.14792}, 2022.

\bibitem[Sitzmann et~al.(2021)Sitzmann, Rezchikov, Freeman, Tenenbaum, and Durand]{lfn}
Vincent Sitzmann, Semon Rezchikov, Bill Freeman, Josh Tenenbaum, and Fredo Durand.
\newblock Light field networks: Neural scene representations with single-evaluation rendering.
\newblock \emph{Advances in Neural Information Processing Systems}, 34:\penalty0 19313--19325, 2021.

\bibitem[Sturm et~al.(2012)Sturm, Engelhard, Endres, Burgard, and Cremers]{ate}
J{\"u}rgen Sturm, Nikolas Engelhard, Felix Endres, Wolfram Burgard, and Daniel Cremers.
\newblock A benchmark for the evaluation of rgb-d slam systems.
\newblock In \emph{2012 IEEE/RSJ international conference on intelligent robots and systems}, pages 573--580. IEEE, 2012.

\bibitem[Team(2024)]{kong2024hunyuanvideo}
Hunyuan Foundation~Model Team.
\newblock Hunyuanvideo: A systematic framework for large video generative models, 2024.

\bibitem[Teed and Deng(2020)]{raft}
Zachary Teed and Jia Deng.
\newblock Raft: Recurrent all-pairs field transforms for optical flow.
\newblock In \emph{Computer Vision--ECCV 2020: 16th European Conference, Glasgow, UK, August 23--28, 2020, Proceedings, Part II 16}, pages 402--419. Springer, 2020.

\bibitem[Unterthiner et~al.(2018)Unterthiner, Van~Steenkiste, Kurach, Marinier, Michalski, and Gelly]{fvd}
Thomas Unterthiner, Sjoerd Van~Steenkiste, Karol Kurach, Raphael Marinier, Marcin Michalski, and Sylvain Gelly.
\newblock Towards accurate generative models of video: A new metric \& challenges.
\newblock \emph{arXiv preprint arXiv:1812.01717}, 2018.

\bibitem[Valevski et~al.(2024)Valevski, Leviathan, Arar, and Fruchter]{gamengen}
Dani Valevski, Yaniv Leviathan, Moab Arar, and Shlomi Fruchter.
\newblock Diffusion models are real-time game engines.
\newblock \emph{arXiv preprint arXiv:2408.14837}, 2024.

\bibitem[Wang et~al.(2024{\natexlab{a}})Wang, Karaev, Rupprecht, and Novotny]{wang2024vggsfm}
Jianyuan Wang, Nikita Karaev, Christian Rupprecht, and David Novotny.
\newblock Vggsfm: Visual geometry grounded deep structure from motion.
\newblock In \emph{Proceedings of the IEEE/CVF conference on computer vision and pattern recognition}, pages 21686--21697, 2024{\natexlab{a}}.

\bibitem[Wang et~al.(2024{\natexlab{b}})Wang, Yuan, Wang, Li, Chen, Xia, Luo, and Shan]{motionctrl}
Zhouxia Wang, Ziyang Yuan, Xintao Wang, Yaowei Li, Tianshui Chen, Menghan Xia, Ping Luo, and Ying Shan.
\newblock Motionctrl: A unified and flexible motion controller for video generation.
\newblock In \emph{ACM SIGGRAPH 2024 Conference Papers}, pages 1--11, 2024{\natexlab{b}}.

\bibitem[Xu et~al.(2024{\natexlab{a}})Xu, Jiang, Huang, Song, Gernoth, Cao, Wang, and Tang]{cavia}
Dejia Xu, Yifan Jiang, Chen Huang, Liangchen Song, Thorsten Gernoth, Liangliang Cao, Zhangyang Wang, and Hao Tang.
\newblock Cavia: Camera-controllable multi-view video diffusion with view-integrated attention.
\newblock \emph{arXiv preprint arXiv:2410.10774}, 2024{\natexlab{a}}.

\bibitem[Xu et~al.(2024{\natexlab{b}})Xu, Nie, Liu, Liu, Kautz, Wang, and Vahdat]{camco}
Dejia Xu, Weili Nie, Chao Liu, Sifei Liu, Jan Kautz, Zhangyang Wang, and Arash Vahdat.
\newblock Camco: Camera-controllable 3d-consistent image-to-video generation.
\newblock \emph{arXiv preprint arXiv:2406.02509}, 2024{\natexlab{b}}.

\bibitem[Xue et~al.(2022)Xue, Hang, Zeng, Sun, Liu, Yang, Fu, and Guo]{HDVILA100M}
Hongwei Xue, Tiankai Hang, Yanhong Zeng, Yuchong Sun, Bei Liu, Huan Yang, Jianlong Fu, and Baining Guo.
\newblock Advancing high-resolution video-language representation with large-scale video transcriptions.
\newblock In \emph{Proceedings of the IEEE/CVF Conference on Computer Vision and Pattern Recognition}, pages 5036--5045, 2022.

\bibitem[Yang et~al.(2024)Yang, Teng, Zheng, Ding, Huang, Xu, Yang, Hong, Zhang, Feng, et~al.]{yang2024cogvideox}
Zhuoyi Yang, Jiayan Teng, Wendi Zheng, Ming Ding, Shiyu Huang, Jiazheng Xu, Yuanming Yang, Wenyi Hong, Xiaohan Zhang, Guanyu Feng, et~al.
\newblock Cogvideox: Text-to-video diffusion models with an expert transformer.
\newblock \emph{arXiv preprint arXiv:2408.06072}, 2024.

\bibitem[Yu et~al.(2025)Yu, Qin, Wang, Wan, Zhang, and Liu]{yu2025gamefactory}
Jiwen Yu, Yiran Qin, Xintao Wang, Pengfei Wan, Di Zhang, and Xihui Liu.
\newblock Gamefactory: Creating new games with generative interactive videos.
\newblock \emph{arXiv preprint arXiv:2501.08325}, 2025.

\bibitem[Yu et~al.(2023)Yu, Lezama, Gundavarapu, Versari, Sohn, Minnen, Cheng, Birodkar, Gupta, Gu, et~al.]{magvit2}
Lijun Yu, Jos{\'e} Lezama, Nitesh~B Gundavarapu, Luca Versari, Kihyuk Sohn, David Minnen, Yong Cheng, Vighnesh Birodkar, Agrim Gupta, Xiuye Gu, et~al.
\newblock Language model beats diffusion--tokenizer is key to visual generation.
\newblock \emph{arXiv preprint arXiv:2310.05737}, 2023.

\bibitem[Zhang et~al.(2025)Zhang, Wang, Xu, Xue, Rupprecht, Zhou, Shen, and Wetzstein]{zhang2025flare}
Shangzhan Zhang, Jianyuan Wang, Yinghao Xu, Nan Xue, Christian Rupprecht, Xiaowei Zhou, Yujun Shen, and Gordon Wetzstein.
\newblock Flare: Feed-forward geometry, appearance and camera estimation from uncalibrated sparse views.
\newblock \emph{arXiv preprint arXiv:2502.12138}, 2025.

\bibitem[Zhou et~al.(2018)Zhou, Tucker, Flynn, Fyffe, and Snavely]{realestate10k}
Tinghui Zhou, Richard Tucker, John Flynn, Graham Fyffe, and Noah Snavely.
\newblock Stereo magnification: Learning view synthesis using multiplane images.
\newblock \emph{arXiv preprint arXiv:1805.09817}, 2018.

\end{thebibliography}
}

\end{document}